\pdfoutput=1

\documentclass[11pt]{article}

\usepackage[]{acl}

\usepackage{times}
\usepackage{latexsym}

\usepackage[T1]{fontenc}

\usepackage[utf8]{inputenc}

\usepackage{microtype}

\usepackage{booktabs}
\usepackage{graphicx}
\usepackage{multirow,makecell,footnote,tablefootnote}
\usepackage{xcolor}

\usepackage{bbold}
\usepackage{url,mathtools,enumitem}
\usepackage[normalem]{ulem}

\definecolor{applegreen}{rgb}{0.55, 0.71, 0.0}

\definecolor{green(pigment)}{rgb}{0.0, 0.65, 0.31}

\title{\textsc{CIns}: Comprehensive Instruction for Few-shot Learning in Task-oriented Dialog Systems}

 \author{
\ \ \ Fei Mi\textsuperscript{†}
\ \ \ Yitong Li\textsuperscript{‡}
\ \ \ Yasheng Wang\textsuperscript{†} 
\ \ \ Xin Jiang\textsuperscript{†}
\ \ \ Qun Liu\textsuperscript{†} \\
\ \ \textsuperscript{†}Huawei Noah's Ark Lab 
\ \ \textsuperscript{‡}Huawei Technologies Co., Ltd.\\
\tt{\{mifei2,liyitong3,wangyasheng,Jiang.Xin,qun.liu\}@huawei.com}
} 

\begin{document}
\maketitle
\begin{abstract}
As labeling cost for different modules in task-oriented dialog (ToD) systems is high, a major challenge in practice is to learn different tasks with the least amount of labeled data.
Recently, prompting methods over pre-trained language models (PLMs) have shown promising results for few-shot learning in ToD. 
To better utilize the power of PLMs, this paper proposes Comprehensive Instruction (\textsc{CIns}) that exploits PLMs with extra task-specific instructions.
We design a schema \emph{(definition, constraint, prompt)} of instructions and their customized realizations for three important downstream tasks in ToD, i.e. intent classification, dialog state tracking, and natural language generation.
A sequence-to-sequence model (T5) is adopted to solve these three tasks in a unified framework.
Extensive experiments are conducted on these ToD tasks in realistic few-shot learning scenarios with small validation data.
Empirical results demonstrate that the proposed \textsc{CIns} approach \emph{consistently} improves techniques that finetune PLMs with raw input or short prompt.
\end{abstract}

\section{Introduction}

Large-scale pre-trained language models (PLMs), such as BERT \cite{devlin2018bert}, UniLM \cite{DBLP:conf/nips/00040WWLWGZH19}, GPT \cite{radford2018improving}, GPT-2 \cite{radford2019language}, T5 \cite{raffel2020exploring} and GPT-3 \cite{brown2020language}, have shown tremendous success in various NLP applications, especially in few-shot or zero-shot learning scenarios.
In task-oriented dialog (ToD) systems, the labeling cost is very high such that the size of well-labeled data is often small. Therefore, few-shot learning in ToD is especially important and valuable in many practical applications.

Many attempts have been proposed to leverage PLMs to improve few-shot learning in ToD. For example, \citet{chen2019bert,chao2019bert,kale2020text} directly finetune a PLM on downstream ToD tasks. 
However, the general objectives and tasks during the model pre-training phase are often very different from the formulation of specific downstream ToD tasks.
To bridge this gap, \citet{kale-rastogi-2020-template,lin2021leveraging} propose to slightly transform the input of downstream ToD tasks to better-matched tasks that PLMs have seen during pre-training.
Such perspective is similar to a recent line of methods, called \textbf{Prompting} or \textbf{Prompt Engineering} \cite{DBLP:conf/eacl/SchickS21,DBLP:conf/naacl/SchickS21,DBLP:conf/acl/GaoFC20,DBLP:journals/corr/abs-2012-11926,DBLP:journals/corr/abs-2107-13586}, to better exploit the capabilities of PLMs.

In Prompting, the input is modified using a ``template'' to form a ``prompt'' to feed to a PLM.
By defining new templates, it unifies the task, objective, and formulation between downstream tasks and pre-training, and Prompting shows strong performance in several few-shot or even zero-shot learning scenarios.
However, one limitation of Prompting is that the ``prompts'' are often \emph{short and concise}~\cite{DBLP:conf/naacl/SchickS21,DBLP:conf/acl/GaoFC20}. We conjecture that the massive amount of information stored in large PLMs might not be adequately exploited with short prompts only. 
Therefore, we are going to study: \emph{can extra instructions further improves short prompts to exploit the few-shot capability of PLMs for ToD?}

To this end, we propose ``\textbf{Comprehensive Instruction} (\textsc{CIns})''. 
Besides a short and concise prompt, we additionally include task-specific \textit{Definition} and \textit{Constraint}. 
\textit{Task Definition} provides a high-level natural language definition and nature of the task itself. \textit{Task Constraint} additionally gives fine-grained task-specific constraint w.r.t output space (e.g. candidate labels, label descriptions, etc.) generated by the PLM.
We formulated the overall schema and task-specific realization of \textsc{CIns} for three downstream tasks (intent classification, dialog state tracking, and natural language generation) in ToD. 
Furthermore, we adopt a Seq2Seq PLM (T5) as a unified framework to solve these three tasks.
In our experiments, we adopt a ``\textit{realistic few-shot learning}'' setting that only uses small validation data with the same size as the few-shot training data.
We contend that this is a more reasonable few-shot setting compared to existing few-shot ToD studies \cite{mi2019meta,peng2020few,kale-rastogi-2020-template,lin2021leveraging} that use full-size validation data. 

The main contribution of this paper is three-fold:
\begin{itemize}[itemsep=-1pt,topsep=0pt,leftmargin=12pt]
\item This is the first attempt to systematically study the effect of add extra task-specific instructions to better exploit pre-trained models for ToD.

\item We propose ``Comprehensive Instruction (\textsc{CIns})'' with a unified schema and task-specific realizations for different ToD tasks. \textsc{CIns} serves as a complement of Prompting to better guide the behavior of powerful PLMs.

\item We conduct extensive experiments on three ToD downstream tasks, including intent classification, dialog state tracking, and natural language generation. A realistic few-shot learning setting is adopted by only utilizing small validation data. Empirical results demonstrate that \textsc{CIns} consistently and notably improves state-of-the-art methods (with or without Prompting) in realistic few-shot learning scenarios.

\end{itemize}

\section{Related Work}

\subsection{Prompting Pre-trained Language Models}

PLMs have shown great success in a number of NLP applications, yet pre-training objectives are often different from downstream tasks.
To bridge this gap. Prompting or Prompt Engineering \cite{DBLP:journals/corr/abs-2107-13586} has been recently studied. In this paper, we focus on ``discrete prompting'' where inputs are wrapped by discrete tokens.
Explorations w.r.t ``continuous prompting'' \cite{DBLP:conf/acl/LiL20,DBLP:journals/corr/abs-2103-10385,DBLP:journals/corr/abs-2104-08691} or how to ensemble multiple prompts \cite{DBLP:conf/eacl/SchickS21,DBLP:conf/naacl/SchickS21,DBLP:journals/tacl/JiangXAN20,DBLP:conf/naacl/QinE21} are beyond the focus of this paper. An overview of related topics can be found at \citet{DBLP:journals/corr/abs-2107-13586}.

Discrete prompting \cite{DBLP:conf/eacl/SchickS21,DBLP:conf/naacl/SchickS21,DBLP:journals/corr/abs-2103-11955,DBLP:conf/acl/GaoFC20,DBLP:journals/corr/abs-2012-11926,DBLP:journals/corr/abs-2101-06804,DBLP:journals/corr/abs-2108-02035} transforms the input to a discrete textual string as a ``prompt'' to feed to a PLM. For classification tasks, a \textit{verbalizer} is often used to map the PLM's output to task labels.
The verbalizer can also be learned for classification tasks \cite{DBLP:conf/acl/GaoFC20,DBLP:journals/corr/abs-2108-02035}.
By defining templates with human intelligence, the few-shot or even zero-shot power of PLMs can be better exploited.
As ``prompts'' are often short and concise, \citet{DBLP:journals/corr/abs-2104-08773} proposed to encode extra task-specific instructions to generalize to new tasks. Our paper is motivated by such idea, while our focus is to study more fine-grained formulations for ToD tasks in realistic few-shot learning settings.

\subsection{Pre-trained Language Models for ToD}

Several large-scale PLMs have been applied to ToD. 
GPT-2 is applied by \citet{budzianowski2019hello,mi2020continual} to
train a response generation model.
\citet{ham2020end,hosseini2020simple,peng2020soloist} proposed to train GPT-2 on different sub-tasks (dialog state tracking, dialog act prediction, and response generation) as a sequence prediction problem. 
BERT is recently applied to different classfication tasks of ToD by \citet{wu2020tod,cai2021slim,mi2021self}.
T5 is also recently applied to ToD by \citet{lin2021leveraging} for dialog state tracking and by \citet{kale-rastogi-2020-template,kale2020text} for natural language generation. 
As GPT-style auto-regressive models are not strong for language understanding tasks, and BERT-style models are not suitable for generation tasks, we adopt Seq2Seq style PLMs, such as T5 or BART \cite{lewis2019bart}, as a unified framework to solve different ToD tasks.

\begin{figure*}[htb!]
    \centering
    \includegraphics[width=0.9\textwidth]{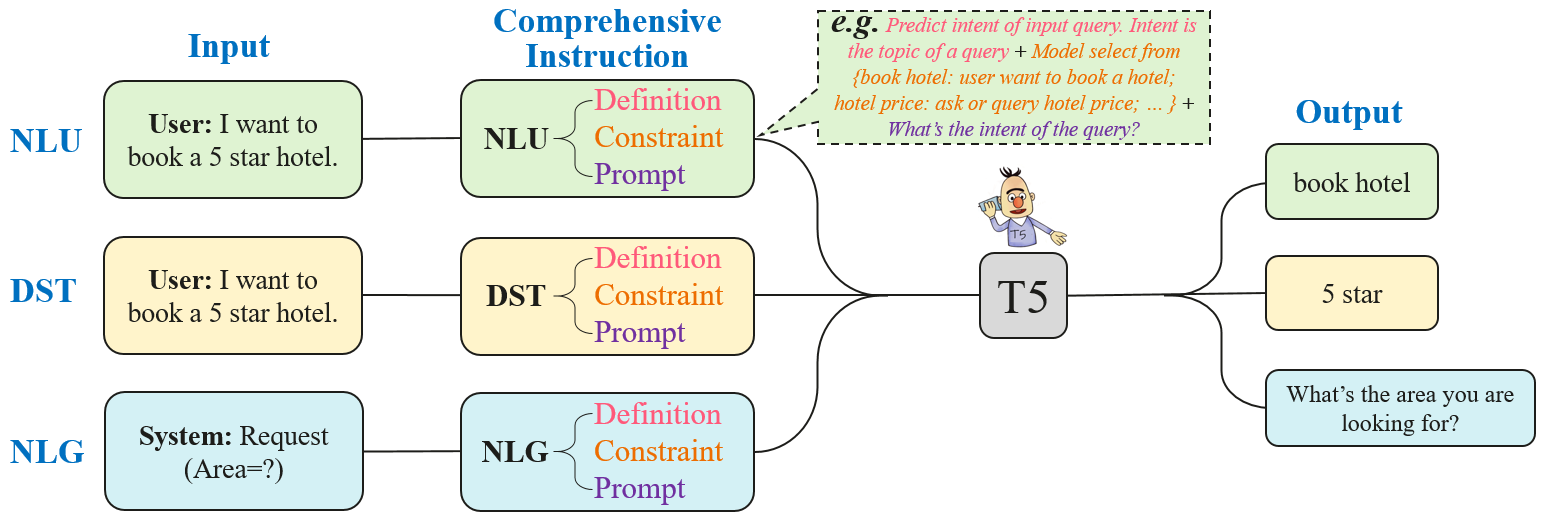}
    \caption{The unified framework of applying Comprehensive Instruction to different ToD downstream tasks. For each task in a row, the input is concatenated with the customized instruction (definition, constraint, prompt) before feeding to a T5 model to generate different types of output. An example instruction for NLU (intent classification) is given in the upper dashed box, and more details for different tasks are elaborated in Table \ref{table:cp_def}.}
    \label{fig:CI_example}
    \vspace{-0.1in}
\end{figure*}

Several studies have also confirmed that PLMs are good few-shot learners for ToD. 
\citet{peng2020soloist} demonstrated few-shot end-to-end response generation from dialog contexts with GPT-2. 
\citet{peng2020few} validated GPT-2 for few-shot natural language generation.
\citet{wu2020tod} showed the effectiveness of BERT for several few-shot classification tasks in ToD.
The few-shot capability of T5 was validated for dialog state tracking \cite{lin2021leveraging} as well as natural language \cite{kale-rastogi-2020-template,kale2020text}. 
The idea of \citet{lin2021leveraging,kale-rastogi-2020-template} bears similarity to Prompting as the input is transformed to better match the pre-trained knowledge of PLMs, and we compared these methods in our paper.

\section{Methodology}

In \S \ref{sec:method1}, we first explain the Seq2Seq (T5) model and how it can unify three downstream tasks (intent classification, dialog state tracking, natural language generation) in ToD. 
In \S \ref{sec:method2}, we explain the unified schema of instructions and how to customize them for different ToD tasks.

\subsection{Seq2Seq (T5) for ToD Tasks}
\label{sec:method1}

We first provide an overview of Seq2Seq model followed by its applications in three tasks in ToD.
As recently proposed by \citet{lewis2019bart,raffel2020exploring}, sequence-to-sequence models unify a variety of natural language processing task, including both classification and generation tasks as:
\begin{equation}
    y = \operatorname{Seq2Seq}(x) = \operatorname{Dec}(\operatorname{Enc}(x)) \, ,
\end{equation}
where both $x$ and $y$ are token sequences.
Input $x$ first goes through a sequence encoder followed by another sequence decoder to produce the output $y$. 
The model is trained used the standard maximum likelihood objective, i.e. using teacher forcing \cite{williams1989learning} and a standard cross-entropy loss. This formulation unifies different ToD tasks, and we elaborate on them later.

There are four common tasks in the ToD pipeline: natural language understanding (NLU), dialog state tracking (DST), dialog management (DM), and natural language generation (NLG).
DM in practice highly depends on business logic to determine suitable system actions. 
Therefore, we focus on NLU, DST, and NLG tasks in this paper.

\paragraph{Intent Classification (IC)} Intent classification is an essential task of NLU in ToD. It predicts the intent label of the user's utterance. For example, the model needs to predict the intent of an utterance ``\textit{I want to book a 5-star hotel}'' as ``\textit{book hotel}''. For IC, the input $x$ to T5 is a user utterance $U_k$, and the output $y$ of T5 is the intent label.

\paragraph{Dialog State Tracking (DST)} Given a dialog history, the task of DST is to predict the value of slots predefined by an ontology.
Following \citet{lin2021leveraging}, the model predicts the value for each (domain, slot) pair.
A dialog history at turn $t$ is a set of alternating utterances between user ($U$) and system ($S$), denoted as $C_t = \{U_1, S_1, \dots, S_{t-1}, U_t\}$.
To predict the value of a slot $i$, the dialog history $C_t$ is concatenated with the name of description $s_i$ of slot $i$.
Then the concatenated sequence $\{C_t, s_i\}$ is fed to the encoder of T5 for the decoder to generate the value $v_i$ of this slot. 

\paragraph{Natural Language Generation (NLG)}

The natural language generation task is to produce a natural language utterance for a \textit{semantic representation} called \textit{dialog
action} produced by the system.
Following \citet{peng2020few,kale2020text}, the ``\textbf{Naive}'' canonical representation $\mathcal{A} = \sum_{i=1}^{A}{a_i(s_i =
v_i)}$ is the set of actions produced
by the system, where $A$ is the total number of actions for this turn. Each action consists of a single intent $a_i$ representing the semantics of the
action, along with optional slot and value pairs $(s_i =
v_i)$. 
For example, 
\begin{equation*}
    \mathcal{A}= [\textit{Inform(name=Rosewood), Inform(star=5)}] \, .
\end{equation*}
However, $\mathcal{A}$ is different from the plain text seen in the pre-trained phase.
To overcome the semantic limitation of $\mathcal{A}$, \citet{kale-rastogi-2020-template} propose to first transform $\mathcal{A}$ to natural language $\mathcal{A}'$ using human-written templates. For example:
\begin{equation*}
    \mathcal{A'}= {\textit{The hotel is called [Rosewood]. It is [5] star.}}
\end{equation*}
The representation of $\mathcal{A'}$ is called Template Guided
Text Generation (\textbf{``T2G2''}, \citet{kale-rastogi-2020-template}), and it achieves strong few-shot NLG performance.
The \textbf{Naive} representation $\mathcal{A}$ or \textbf{T2G2} $\mathcal{A'}$ is feed to the encoder of T5, and the decoder generates a natural language utterance as a response. 

Examples of T5 for IC, DST, and NLG tasks are illustrated in Figure \ref{fig:CI_example} without looking at the middle column (``Comprehensive Instruction'').

\definecolor{brightube}{rgb}{0.82, 0.62, 0.91}
\definecolor{orangepeel}{rgb}{1.0, 0.62, 0.0}
\definecolor{pinksherbet}{rgb}{0.97, 0.56, 0.65}
\definecolor{blue-violet}{rgb}{0.54, 0.17, 0.89}
\begin{table*}[htb!]
\small
\setlength{\tabcolsep}{5pt}
    \centering
    \begin{tabular}{c|p{4.5cm}|p{5.5cm}|p{3.5cm}}
        \toprule
         & \textcolor{pinksherbet}{\textbf{\normalsize{Task Definition}}} & \textcolor{orangepeel}{\textbf{\normalsize{Task Constraint}}} & \textcolor{blue-violet}{\textbf{\normalsize{Prompt}}} \\
        \midrule
        \multirowcell{3}{\textbf{IC}} & Predict the intent of the input query.  Intent is the main topic or purpose of a query. & Model needs to select the most suitable intent from: \{candidate labels + label descriptions\}$^\clubsuit$. & What is the intent of the given query? \\
        \midrule
        \multirowcell{3}{\textbf{DST}} & Predict the \{slot description\}$^\spadesuit$ requested by User. & Select the most suitable value from: \{candidate values\}$^\diamondsuit$. If multiple values appear, select the latest one. & What is / Whether the \{slot description\}?\\
        \midrule
        \multirowcell{3}{\textbf{NLG}$^\heartsuit$} & Verbalize the input representation / Paraphrase the input sentences. & The output should be natural and concise, and it should preserve the meaning and information of the input. & How to verbalize the input? / What is the paraphrase utterance of the input? \\
        \bottomrule
    \end{tabular}
    \caption{$^\clubsuit$ this is the concatenation of all candidate intents with its corresponding descriptions. $^\spadesuit$ two types of descriptions for the slot ``hotel-stars'' can be founded in Table \ref{table:prompt_type}. $^\diamondsuit$ candidate values for categorical slots (e.g. area, type), and it is left empty for open non-categorical slots (e.g. time, name). $^\heartsuit$ Two types of task definitions and prompts are used for Naive ($\mathcal{A}$) and T2G2 ($\mathcal{A}'$) representations mentioned in Section \ref{sec:method1} respectively.}
    \vspace{-0.12in}
    \label{table:cp_def}
\end{table*}

\subsection{Comprehensive Instruction for ToD}
\label{sec:method2}

This section first explains two existing types (Standard, Prompt Engineering) of input to Seq2Seq. Then, we explain how to formulate the proposed method (Comprehensive Instruction), and how to design it for different downstream tasks in ToD.

\paragraph{Standard (STD)} The standard input to Seq2Seq models is the raw set of input tokens. We explained different kinds of standard input to T5 in \S \ref{sec:method1}. For example, the raw token sequence \textit{``I want to book a 5-star hotel''} serves as the input for intent classification to predict the label ``book hotel''.

\paragraph{Prompt Engineering (PE)} To better utilize the capability of PLMs, PE constructs task-specific \textit{prompts} around the raw input before feeding to PLMs. For example, \textit{```I want to book a 5-star hotel' \textcolor{blue}{\uline{What does the previous query ask about?}}''}. Underlined tokens in blue are human-designed prompt tokens to help the PLM understand the task.
The idea of PE is to find a proper prompt for the task to bridge the gap between the downstream task and the PLM's capability. 
PE shows promising results in various few-shot learning applications.

\subsubsection{Schema of Comprehensive Instruction}
However, prompts in PE are often concise. To fully exploit the capability of PLMs, we propose to construct Comprehensive Instructions (\textsc{CIns}) on top of PE. The idea is to provide extra task-specific instructions for PLMs to understand critical \textit{abilities} to solve the task. Besides the short prompts, we propose to add \textbf{task definition} and \textbf{task constraints} as instructions. An abstract configuration of \textsc{CIns} compared to PE and STD can be visualized in Figure~\ref{fig:CI_compare}. The goal of these two additional components are elaborated below:

\begin{figure}[t!]
    \centering
    \includegraphics[width=0.47\textwidth]{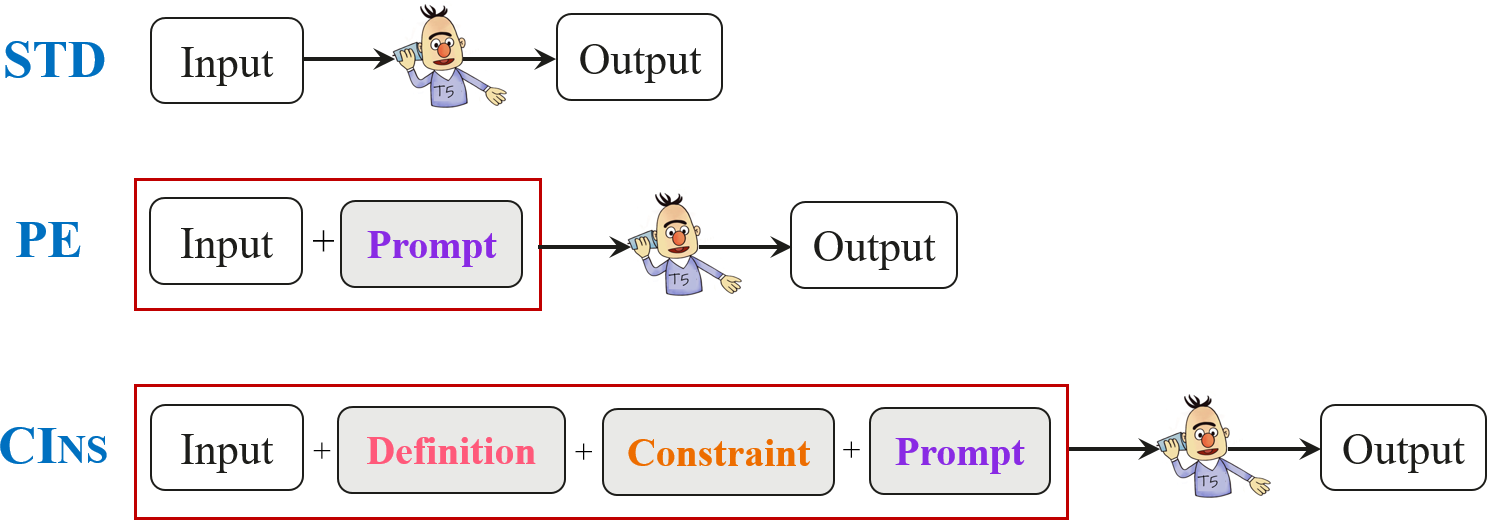}
    \caption{Formulation comparison for Standard input, Prompt Engineering, and Comprehensive Instruction.}
    \label{fig:CI_compare}
    \vspace{-0.1in}
\end{figure}

\begin{itemize}[itemsep=-1pt,topsep=2pt,leftmargin=12pt]
\item \textbf{Task Definition}: it provides a high-level natural language definition of the task itself. It describes the nature of the task, such as what types of input and output that the PLM is dealing with.

\item \textbf{Task Constraint}: it gives more fine-grained task-specific constraint w.r.t output generated by the PLM. For example, the candidate labels and the descriptions of each candidate label. Task constraints aim to compliments the task definition to give more instructions about what the should the PLM output.
It is not independent of the task definition, and it put more emphasis on the constraints w.r.t. the output space.
\end{itemize}

Different components of \textsc{CIns} are concatenated using a [SEP] token, and each component starts with a leading identifier (\textit{``Input:'', ``Definition:'', ``Constraint:'', ``Prompt:''}).


\subsubsection{\textsc{CIns} for ToD Tasks}
\label{sec:ci_tod}
This section elaborates the realization of Comprehensive Instruction for three ToD tasks summarized in Table~\ref{table:cp_def}. We formulate the ``prompt'' using ``Question'' expressions, and the advantage over ``Declarative'' expressions will be analyzed in \S~\ref{sec:analysis}.
Next, we mainly explain ``task definition'' and ``task constraint'' for different tasks.

\paragraph{\textsc{CIns} for IC} For intent classification, the definition explains the task followed by the meaning of ``intent''. 
To add constraints w.r.t. the output space, we include candidate intent labels and their corresponding descriptions in the constraint component. 
More specifically, for $K$ candidate intents in a domain, we concatenate all intent names $n_i$ with their descriptions $d_i$ as $\{n_1: d_1, \dots, n_1: d_K\}$ to add to the constraint.
The exact intent descriptions are provided in Appendix \ref{appendix:ic_description}.

\paragraph{\textsc{CIns} for DST} For dialog state tracking, the model predicts the value of different slots requested by the user from a dialog context. This task definition is encoded in our task definition component. ``\{slot description\}'' follows \citet{lin2021leveraging} that encodes the type of a slot.
In the task constraint component, we encode the constraint of candidate values for slots (e.g. \textit{are, type, price-range}, etc.) that have categorical values \cite{zhang2020find,rastogi2020towards}. For non-categorical slots (\textit{name, time}) that have open values, no such constraint is enforced. Furthermore, a slot might be mentioned multiple times in a dialog history, and the correct value of interest often needs to be captured in its latest mention, and we also encode this information in the task constraint.
Lastly, for slots (e.g. \textit{has-internet, has-parking}) with ``\textit{yes/no}'' values, we formulate the prompt question with ``\textit{Whether}'', otherwise, the prompt question starts with ``\textit{What}''.

\paragraph{\textsc{CIns} for NLG} For natural language generation, the task definition is customized for two types of input representations (c.f. \S \ref{sec:method1}). For ``\textbf{Naive}'' representation, the task is to \textit{verbalize} the semantic representation to a natural language utterance. For ``\textbf{T2G2}'' representation, \textit{paraphrasing} is a more precise definition. Exact task definitions for these two cases are given in Table~\ref{table:cp_def}. In terms of the constraint, we tell the model the output utterance should be \textit{natural and concise}, and it should also \textit{preserve the meaning and information} of the original input representation for the fidelity concern.

\definecolor{green(html/cssgreen)}{rgb}{0.0, 0.5, 0.0}
\section{Experiment}

\begin{table*}[htb!]
\centering
\small
\setlength{\tabcolsep}{1.3pt}
\makebox[0pt][c]{\parbox{1\textwidth}{%
    \begin{minipage}[b]{0.32\hsize}\centering
        \begin{tabular}{c|c|cccccc}
        \toprule
          \multicolumn{1}{c}{\textbf{T5-small}}& \multicolumn{1}{c}{} & \textbf{Average} & \textbf{Bank} & \textbf{Home} & \textbf{Travel}  & \textbf{Utility} & \textbf{Auto} \\
         \midrule
          \multirow{3}{*}{1-shot} & STD & 35.5 $\pm$ 1.9 & 32.2  & 29.0 & 25.0 & 47.6 & 43.6 \\
          & PE &  41.7 $\pm$ 2.9 & 38.0  & 34.4 & 33.6 & 54.4 & 48.1\\
          & \textsc{CIns} &  \textbf{71.1} $\pm$ 4.4 & \textbf{62.6}  & \textbf{57.9} & \textbf{80.4}  & \textbf{84.2} & \textbf{70.2} \\ 
          \midrule
          \multirow{3}{*}{5-shot} & STD &  72.4 $\pm$ 1.7 & 66.2  & 64.4 & 74.5  & 80.2 & 76.5  \\
          & PE &  76.8 $\pm$ 2.5 & 71.9  & 65.8 & 83.0  & 84.1 & 79.5 \\ 
          & \textsc{CIns} &  \textbf{85.6} $\pm$ 1.7 & \textbf{82.4}  & \textbf{71.3} & \textbf{94.5}  & \textbf{93.3} & \textbf{86.2} \\ 
          \midrule
          Full & STD & 96.8 & 93.6 & 96.0 & 98.0 & 98.4  & 98.0 \\
        \bottomrule
    \end{tabular}
    \end{minipage} \hspace{85pt}
    \begin{minipage}[b]{0.32\hsize}\centering
        \begin{tabular}{c|c|cccccc}
        \toprule
         \multicolumn{1}{c}{\textbf{T5-base}} & \multicolumn{1}{c}{}& \textbf{Average} & \textbf{Bank} & \textbf{Home} & \textbf{Travel} & \textbf{Utility} & \textbf{Auto} \\
         \midrule
          \multirow{3}{*}{1-shot} & STD & 56.0 $\pm$ 3.6 & 56.6  & 41.6 & 58.5  & 62.7 & 60.5  \\
          & PE &  61.4 $\pm$ 3.1 & 62.3  & 48.7 & 59.2  & 74.9 & 62.0 \\
          & \textsc{CIns} &  \textbf{79.2} $\pm$ 2.2 & \textbf{80.8}  & \textbf{60.2} & \textbf{87.3}  & \textbf{86.2} & \textbf{81.5} \\ 
          \midrule
          \multirow{3}{*}{5-shot} & STD &  85.8 $\pm$ 2.1 & 83.3  & 72.1 & 91.1  & 93.8 & 89.0  \\
          & PE &  87.0 $\pm$ 1.3 & 86.7  & 72.2 & 92.4  & 94.8 & 89.0  \\ 
          & \textsc{CIns} &  \textbf{91.1} $\pm$ 2.2 & \textbf{89.1}  & \textbf{80.2} & \textbf{97.1}  & \textbf{95.4} & \textbf{93.7} \\ 
          \midrule
          Full & STd & 97.4 & 94.7 & 96.7 & 98.1 & 98.7 & 98.5 \\
        \bottomrule
    \end{tabular}
    \end{minipage}
}}
\caption{Accuracy in percentage [\%] for intent classification task with T5-small (left) and T5-base (right). The ``Average'' column reports the average results and the standard deviations of 5 domains.}
\label{table:nlu}
\end{table*}
\begin{table*}[htb!]
\centering
\small
\setlength{\tabcolsep}{2pt}
\makebox[0pt][c]{\parbox{1\textwidth}{%
    \begin{minipage}[b]{0.32\hsize}\centering
        \begin{tabular}{c|c|cccccc}
        \toprule
         \multicolumn{1}{c}{\textbf{T5-small}} & \multicolumn{1}{c}{} & \textbf{Average} & \textbf{Attr.} & \textbf{Hotel} & \textbf{Rest.} & \textbf{Taxi} & \textbf{Train} \\
         \midrule
          \multirow{3}{*}{1\% Data} & STD & 33.6 $\pm$ 2.3 & 25.1  & 24.4 & 32.9 & 60.0 & 25.4\\
          & PE &  41.7 $\pm$ 3.6 & 36.0  & 25.9 & \textbf{33.8} & 59.9 &  52.3\\
          & \textsc{CIns} &  \textbf{43.1} $\pm$ 1.8 & \textbf{42.0}  & \textbf{27.3} & 32.7 & \textbf{60.4} & \textbf{53.1}\\ 
          \midrule
          \multirow{3}{*}{5\% Data} & STD &  55.1 $\pm$ 2.2 & 54.3  & 43.0 & 50.2 & 59.0 & 69.1 \\
          & PE &  55.7 $\pm$ 1.6 & \textbf{57.0}  & 42.2 & 51.1 & 59.2 & 68.7\\ 
          & \textsc{CIns} &  \textbf{57.0} $\pm$ 1.1 & 56.9  & \textbf{43.4} & \textbf{51.3} & \textbf{61.8} & \textbf{71.3}\\ 
          \midrule
          Full & STD & 72.0 & 71.3 & 59.5 & 68.0 & 81.4 & 80.0\\
        \bottomrule
    \end{tabular}
    \end{minipage} \hspace{85pt}
    \begin{minipage}[b]{0.32\hsize}\centering
        \begin{tabular}{c|c|cccccc}
        \toprule
         \multicolumn{1}{c}{\textbf{T5-base}} & \multicolumn{1}{c}{} & \textbf{Average} & \textbf{Attr.} & \textbf{Hotel} & \textbf{Rest.} & \textbf{Taxi} & \textbf{Train} \\
         \midrule
         \multirow{3}{*}{1\% Data} & STD & 45.5 $\pm$ 3.5 & 41.5 & 30.8 & 36.7 & 58.4 & 60.1 \\
         & PE &  46.5 $\pm$ 3.1 & 41.7  & 31.3 & 39.0 & 59.5 & \textbf{61.2} \\
         & \textsc{CIns} &  \textbf{47.9} $\pm$ 2.1 & \textbf{45.6}  & \textbf{33.9} & \textbf{40.6} & \textbf{59.7} & 60.3\ \\ 
          \midrule
         \multirow{3}{*}{5\% Data} & STD & 58.8 $\pm$ 1.7 & 59.7 & 44.5  & \textbf{54.1}  & 62.8  &  73.2 \\
         & PE & 57.8 $\pm$ 2.9 & 59.3 & 43.7 & 51.9 & 61.7 & 72.6\\
         & \textsc{CIns} & \textbf{59.7} $\pm$ 2.4 & \textbf{61.2} & \textbf{46.2}  & 53.9 & \textbf{63.3} & \textbf{73.8} \\
         \midrule
          Full & STD & 72.8 & 73.6 & 60.5 & 66.4 & 81.9 &  81.8\\
        \bottomrule
    \end{tabular}
    \end{minipage}
}}
\caption{Joint Goal Accuracy in percentage [\%] for few-shot dialog state tracking using T5-small (left) and T5-base (right). The ``Average'' column reports the average results and the standard deviations of 5 domains.}
\vspace{-0.1in}
\label{table:dst}
\end{table*}

\subsection{Few-shot Datasets}

We evaluate three different ToD downstream tasks with three different datasets respectively.

\paragraph{OOS} For intent classification, we use a benchmark dataset from \citet{larson2019evaluation}. Apart from the single out-of-scope intent, it contains 150 intents in 15 domains.
Each domain contains 15 intents with 1,500/300/450 instances for train/validation/test, and data are balanced across different intents.
Several domains are similar to each other, and we test on 5 representative domains (\textit{Bank}, \textit{Home}, \textit{Travel}, \textit{Utility}, and \textit{Auto}).
For few-shot setups, we sample $k$ instances per intent from the training data, noted as ``$k$-shot''.

\paragraph{MultiWOZ2.0} We evaluate dialog state tracking task using MultiWOZ2.0 \cite{budzianowski2018multiwoz}. It contains 8,420/1,000/1,000 dialogues for train/validation/test spanning over 7 domains.
Following \citet{wu2019transferable,lin2021leveraging}, we adopt \textit{attraction}, \textit{hotel}, \textit{restaurant}, \textit{train}, and \textit{taxi} domains for training, as the test set only contains these 5 domains. In few-shot setups, we experiment with ``k\% Data'', i.e. only k\% of the training dialogs are used.

\paragraph{FewShotSGD} \citet{kale-rastogi-2020-template} is the version of the schema-guided-dataset \cite{rastogi2019scalable} for natural language generation. 
The full train/validation/test sets contain 160k/24k/42k utterances.
In ``$k$-shot'' experiments, $k$ dialogs from each 14 training domains are sampled from the training data.
We use the same 5/10-shot training split as in \citet{kale-rastogi-2020-template} because they both contain utterances for every dialog act and slot present in the full training set.\footnote{1-shot training data is not provided with such property}

To test ``realistic few-shot learning'' scenarios mentioned before, we down-sample validation data to be the same size as the few-shot training data in all our experiments.
An analysis of validation data size is included in Appendix~\ref{appendix:val_size}. The exact data sizes in different tasks and configurations are elaborated in Appendix \ref{appendix:data}. 

\subsection{Experiment Settings}
\label{sec:exp_setting}

We tested \textbf{T5-small} (60M parameters, 6
encoder-decoder layers) as well as \textbf{T5-base} (110M parameters, 12
encoder-decoder layers) using the huggingface repository.\footnote{\url{https://huggingface.co}}
All models are trained using
AdamW (Loshchilov and Hutter, 2018) optimizer
with the initial learning rate of 1e-4 for DST and NLG, and 3e-4 for IC. 
In all experiments, we train the models with batch size 8 for 30 epochs for IC, 20 epochs for DST, and 50 epochs for NLG.
Early stop according to the loss on the validation set. 
In the testing phase, we use greedy decoding.
We use 4 NVIDIA V100 GPUs for all of our experiments.

For comparison, we consider two baselines, STD and PE.
For prompting based method, both PE and \textsc{CIns}, six prompts are tested, 
and we report the results with the best prompt choice without mentioned specifically.
For STD, we also report a \textit{upper bound} using all labeled training data and validation data, referred to as ``Full''.
For all few-shot experiments, we report mean and standard deviation with three different random seeds to reduce training/validation data sampling variance.


\subsection{Main Experiment Results}
\label{sec:main_exp}

\begin{figure*}[htb!]
    \centering
    \includegraphics[width=0.97\textwidth]{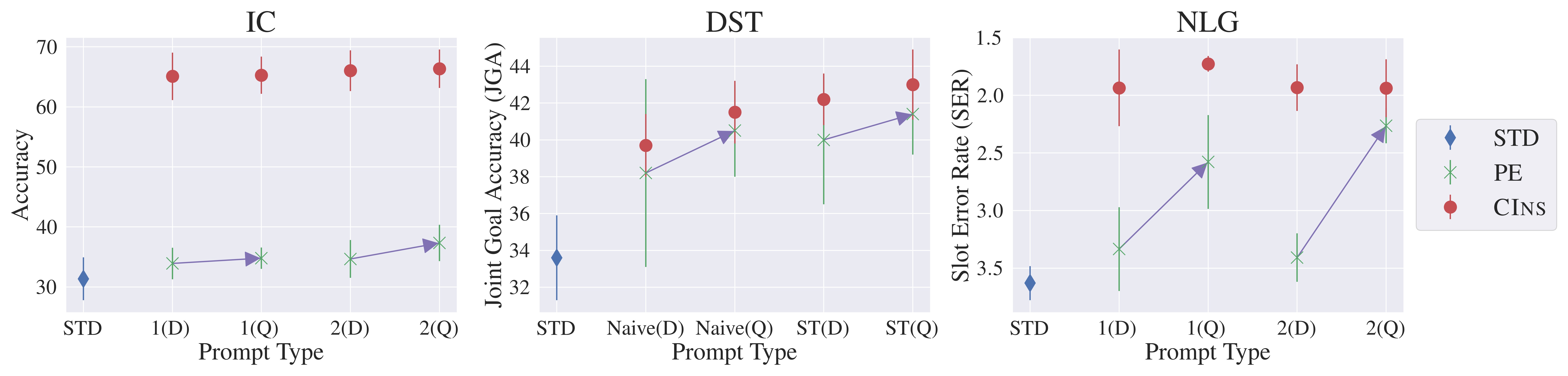}
    \vspace{-0.1in}
    \caption{Results of IC (1-shot), DST (1\% data), NLG (1-shot) with different types of prompts with T5-small. Different prompts tested for IC and DST are provided in Table \ref{table:prompt_type}. For the same prompt backbone, the version with \textbf{(D)} stands for a \textbf{Declarative} expression, while the version with \textbf{(Q)} stands for a \textbf{Question} expression. Standard deviations over different random seeds are also plotted.}
    \vspace{-0.05in}
    \label{fig:prompt_type}
\end{figure*}

\paragraph{Intent Classification}
Accuracy of few-shot intent classification on 5 domains over OOS is presented in Table \ref{table:nlu}.
For both T5-small and T5-base, 1-shot and 5-shot settings are considered, and the column headed with ``Average'' averages results of 5 domains.
We see that STD performs the worst in different configurations and that PE consistently outperforms STD.
\textsc{CIns} significantly outperforms both STD and PE in all configurations, especially with fewer label (1-shot).
In 1-shot setting, \textsc{CIns} achieves 29.4\% and 17.8\% higher average accuracy than PE for T5-small and T5-base respectively.
In 5-shot settings, the above two margins are 8.8\% and 4.1\%.
These results demonstrate that \textsc{CIns} effectively boost the few-shot learning capability for intent classification.

\paragraph{Dialog State Tracking}
Results of DST on MultiWOZ2.0 are presented in Table~\ref{table:dst}. 1\% and 5\% labeled data are tested w.r.t. T5-small and T5-base.
The common evaluation metrics joint goal accuracy (JGA) \cite{budzianowski2018multiwoz,wu2019transferable} is used, which checks whether the predicted states (domain, slot, value) match the ground truth states given a context. 
PE on average performs better than STD except for the configuration with 5\% labeled data using T5-base. 
We  see that \textsc{CIns} consistently improves the averaged JGA over both STD and
PE in different configurations.
For example, \textsc{CIns} has 3.1\% and 9.5\% average JGA improvement over PE and STD respectively when 1\% labeled data are used with T5-small. Smaller margins can be observed for other configurations.

\begin{table}[t!]
    \centering
    \small 
    \setlength{\tabcolsep}{5pt}
    \begin{tabular}{cc|c|cc}
        \toprule
        \textbf{T5-small} & \multicolumn{1}{c}{}& \multicolumn{1}{c}{} & \textbf{SER} $\downarrow$ & \textbf{BLEU} $\uparrow$ \\
         \midrule
          \multirow{7}{*}{Naive} & \multirow{3}{*}{5-shot} & STD & 10.2 $\pm$ 0.49  & 17.3 $\pm$ 0.23 \\
          & & PE &  9.8 $\pm$ 0.43 & 17.1 $\pm$ 0.17  \\
          & & \textsc{CIns} & \textbf{6.9} $\pm$ 0.25  & \textbf{17.5} $\pm$ 0.13  \\ 
        \cmidrule{2-5}
        & \multirow{3}{*}{10-shot} & STD &  5.9 $\pm$ 0.30  & 19.0  $\pm$ 0.18  \\
          & & PE &  5.2 $\pm$ 0.34  & 19.2 $\pm$ 0.26   \\
          & & \textsc{CIns} &  \textbf{4.6} $\pm$ 0.28   & \textbf{19.4} $\pm$ 0.07    \\ 
          \cmidrule{2-5}
          & Full & STD & 1.0 & 26.3 \\
        \midrule
         \multirow{7}{*}{T2G2} & \multirow{3}{*}{5-shot} & STD & 3.6 $\pm$ 0.15  & 25.5 $\pm$ 0.05 \\
          & & PE &  2.3 $\pm$ 0.15  & 26.0 $\pm$ 0.05  \\
          & & \textsc{CIns} & \textbf{1.7} $\pm$ 0.07  & \textbf{26.3} $\pm$ 0.11  \\ 
          \cmidrule{2-5}
        & \multirow{3}{*}{10-shot} & STD &  3.5 $\pm$ 0.15  & 25.9  $\pm$ 0.07  \\
          & & PE &  1.9 $\pm$ 0.33  & 26.4 $\pm$ 0.12   \\
          & & \textsc{CIns} &  \textbf{1.3} $\pm$ 0.07   & \textbf{26.7} $\pm$ 0.04    \\ 
          \cmidrule{2-5}
          & Full & STD & 0.4 & 28.6 \\
        \bottomrule
    \end{tabular}
    \caption{\normalsize Performance of few-shot natural language generation using T5-small as the generation model. Two types of semantic representations are tested: Naive and T2G2.}
    \label{table:nlg}
    \vspace{-0.1in}
\end{table}

\definecolor{brightube}{rgb}{0.82, 0.62, 0.91}
\definecolor{pinksherbet}{rgb}{0.97, 0.56, 0.65}
\definecolor{asparagus}{rgb}{0.53, 0.66, 0.42}
\definecolor{beaublue}{rgb}{0.74, 0.83, 0.9}
\definecolor{applegreen}{rgb}{0.55, 0.71, 0.0}

\begin{savenotes}
\begin{table*}[htb!]
\small
\setlength{\tabcolsep}{5pt}
    \centering
    \begin{tabular}{c|c|c|c}
        \toprule
         & \textbf{Prompt} & \textbf{Declarative (D)} & \textbf{Question (Q)} \\
        \midrule
        \multirow{2}{*}{\textbf{IC}} & 1 & The \textcolor{applegreen}{given query asks about}: & Question: What does the \textcolor{applegreen}{given query ask about}? \\
        & 2 & The \textcolor{pinksherbet}{intent of the given query} is: & Question: What is the \textcolor{pinksherbet}{intent of the given query}? \\
        \midrule
       \multirowcell{2}{\textbf{DST} \\ (hotel\_stars)} & Naive$^\dag$ & \textcolor{applegreen}{stars of the hotel} & Question: What is the \textcolor{applegreen}{stars of the hotel}? \\
         & Slot Type (ST)$^\ddag$  & \textcolor{pinksherbet}{number of stars of the hotel} & Question: What is the \textcolor{pinksherbet}{number of stars of the hotel}? \\
         \midrule
          \multirowcell{2}{\textbf{NLG}} & 1 & \textcolor{applegreen}{Paraphrase the input:} & Question: what is the \textcolor{applegreen}{paraphrase of the input}? \\
         & 2  & \textcolor{pinksherbet}{Rewrite the input:} &  Question: what is the \textcolor{pinksherbet}{rewriting of the input}? \\
        
        \bottomrule
    \end{tabular}
    \caption{Different prompts for IC, DST and NLG (``T2G2''). In each row, two expressions (declarative and question) are considered for a highlighted prompt root. For DST, we use the slot ``hotel-stars'' as an example, and more descriptions can be founded in \citet{lin2021leveraging}. $^\dag$ Transform "domain-slot" to "[slot] of the [domain]. $^\ddag$ Transform "domain-slot" to "[slot type] [slot] of the [domain]" \cite{lin2021leveraging}.}
    \label{table:prompt_type}
    \vspace{-0.1in}
\end{table*}
\end{savenotes}

\paragraph{Natural Language Generation}
Results of T5-small for 5-shot and 10-shot NLG using two types of semantic representations ``Naive'' and ``T2G2'' are included in Table~\ref{table:nlg}.
We only report T5-small as it already performs well enough.
Following prior works \cite{wen2015semantically,kale-rastogi-2020-template}, we use
BLEU \cite{papineni2002bleu} and Slot Error Rate (SER \citet{duvsek2019neural}) as metrics.
SER measures the fraction of generated texts where at least one slot was not correctly copied from the structured data.
\textsc{CIns} outperforms both PE and STD in all configurations and metrics with notable margins. Compared to PE, \textsc{CIns} improves SER of ``Naive'' by 2.9\% and 0.6\% in 5-shot and 10-shot settings respectively. These two margins are 0.6\% and 0.6\% for ``T2G2''. The improvement margins over STD are larger. Moreover, when ``T2G2'' is used as the input representation, \textsc{CIns} with only 5-shot or 10-shot training samples achieves comparable performance compared to ``Full''.

Altogether, our experiments on three different downstream tasks reveal that:
\begin{itemize}[itemsep=-1pt,topsep=2pt,leftmargin=12pt]
    \item Comprehensive Instruction provides complimentary benefits over standard input and prompting. \textsc{CIns} consistently improves both PE and STD in all configurations of three few-shot ToD tasks. The margin is evident on IC, indicated by 17-30\% average gain over PE with 1-shot data; 4-9\% gain with 5-shot data. The margin is smaller on two more challenging DST and NLG tasks.
    \item Comprehensive Instruction bridges the gap between few-shot learning and full supervision. STD and PE with few-shot labeled data perform much worse than models trained with all labeled data (“Full”) for IC and NLG. \textsc{CIns} largely improves performances on these two tasks with results comparable to “Full”.

\end{itemize}

\subsection{Analysis and Discussions}

\label{sec:analysis}
\paragraph{What is a good prompt?}
In this experiment, we study what makes it a good prompt with T5 used as the backbone model. In Table~\ref{table:prompt_type}, we present two types of ``prompt roots'' highlighted in different colors for IC, DST and NLG (``T2G2''). For each prompt root (row), two expressions are considered in declarative and question forms.\footnote{A prefix ``Question:'' is adopted as we found that adding it achieves slightly better performance because this prefix is commonly seen during the T5 pre-training phase.}
In Figure \ref{fig:prompt_type}, we present results of using these different prompts with T5-small for IC (1-shot), DST (1\% Labeled Data) and NLG (5-shot). 
Results for PE are illustrated by green points. For the same prompt root, prompts with a question (Q) expressions always outperform prompt with declarative (D) expressions for both PE and \textsc{CIns}.
Such a pattern is more obvious for PE, and it can be better visualized by purple arrows.

\paragraph{Does \textsc{CIns} improve different prompts?}
In this experiment, we study whether \textsc{CIns} achieves consistent improvements with different prompts. In Figure~\ref{fig:prompt_type}, we compare \textcolor{red}{\textsc{CIns}} and \textcolor{green(html/cssgreen)}{PE} with four prompts mentioned in Table~\ref{table:prompt_type}. For each prompt, regardless of declarative or questions expressions, \textsc{CIns} outperforms PE. This results validates that adding task-specific definition and constraint as instructions is beneficial for prompts in PE.

\begin{table}[t!]
    \centering
    \begin{tabular}{l|ccc}
        \toprule
          \textbf{T5-small} & \textbf{IC} & \textbf{DST} & \textbf{NLG}  \\
         \midrule
           \textsc{CIns} & 71.1 & 43.1 & 1.7 \\
            { } w/o Description & 65.6 & 41.2  & - \\
            { } w/o Definition & 69.8 & 42.5 & 1.9 \\ 
            { } w/o Prompt & 70.1 & 42.2 & 2.3  \\
            { } w/o Constraint & 45.7 & 40.8  &  2.4 \\
            PE & 41.7 & 40.0 & 2.6 \\
        \bottomrule
    \end{tabular}
    \caption{Ablation Study for \textsc{CIns} for IC (1-shot), DST (1\% data), NLG (5-shot SER) with T5-small. ``Description'' stands for IC label and DST slot descriptions. }
    \label{table:ablation}
    \vspace{-0.14in}
\end{table}

\paragraph{Is \textsc{CIns} robust?}
For main experiments conducted in \S~\ref{sec:main_exp}, we experiment with different data sizes and model sizes for IC and DST; different data sizes and input forms for NLG. In total, we have 12 configurations (4 for IC, 4 for DST, 4 SERs for NLG). For each configuration, standard deviations of three random seeds are also reported. We could see that \textsc{CIns} achieves the lowest standard deviation in 9/12 configurations. This result demonstrates that \textsc{CIns} is more robust to the data sampling variance in few-shot learning settings.
Furthermore, we could see from Figure \ref{fig:prompt_type} that the performance of \textsc{CIns} is also less sensitive to the prompts than PE. For example, the performance of 2(D) vs. 2(Q) or ST(D) vs. ST(Q) differs a lot for PE, while they perform similarly for \textsc{CIns}. 
Therefore, We contend that extra task-specific instructions in \textsc{CIns} improve model robustness w.r.t. the choice of few-shot training data as well as prompt.

\paragraph{Ablation study}

In Table \ref{table:ablation}, we compare several simplified versions of \textsc{CIns} to understand the effects of different components. ``w/o Definition'', ``w/o Constraint'', and ``w/o Prompt'' are intuitive. ``w/o Description'' for IC removes label descriptions in constraint, and ``w/o Description'' for DST replace the slot description from ``Slot Type'' to ``Naive'' \cite{lin2021leveraging}.
We observe that: (i) label and slot descriptions are beneficial. Removing it degrades performance by 5.5\% and 1.9\% on IC and DST respectively. (ii) Task definition and Prompt are both concise but advantageous. Dropping either of them (``w/o Definition'', ``w/o Prompt'') hurts performance slightly. (iii) Task-specific constraint is a critical component, indicated by relatively large performance drop on three tasks when removing it (``w/o Constraint''). Nevertheless, it still outperforms ``PE'', meaning that the model still learns from task definitions.

\begin{figure}[t!]
    \centering
    \includegraphics[width=0.42\textwidth]{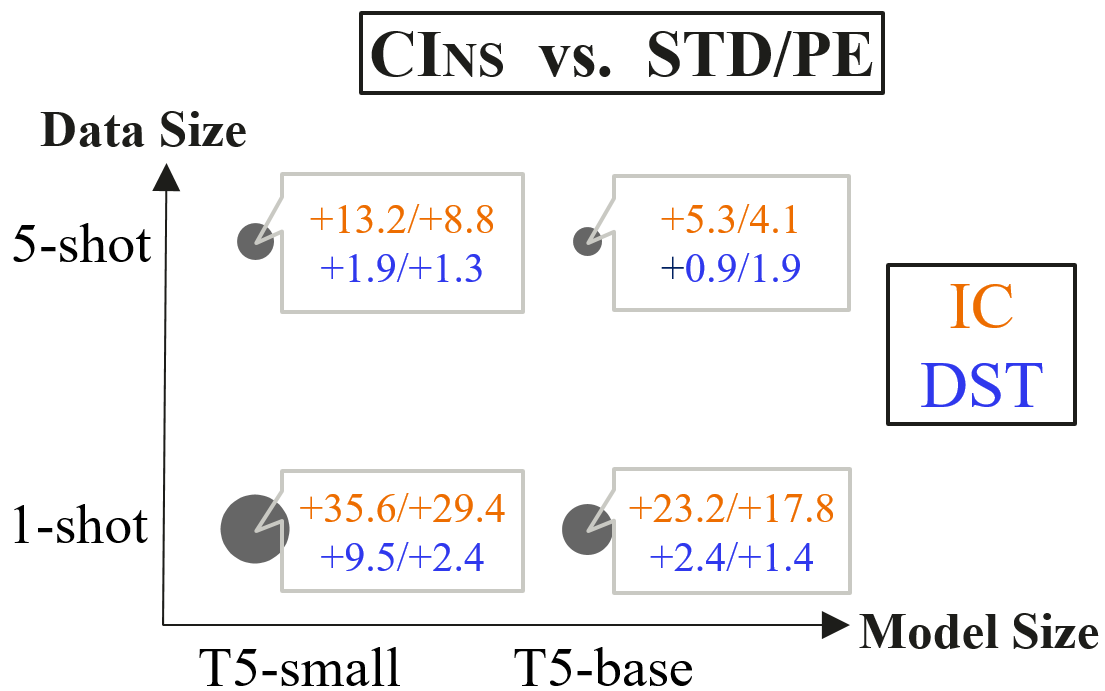}
    \caption{Accuracy improvements of \textsc{CIns} over STD/PE in different configurations of intent classification (\textcolor{orange}{orange}) and dialog state tracking (\textcolor{blue}{blue}) tasks.}
    \label{fig:data_model_size}
    \vspace{-0.1in}
\end{figure}

\paragraph{Effect of Different Model and Data Sizes}
\label{appendix:model_data_size}

In Figure \ref{fig:data_model_size}, we plot the improvements of \textsc{CIns} over STD and PE in different configurations of IC (red) and DST (blue).
We could see that the improvement margin of \textsc{CIns} over STD/PE is largest in the 1-shot setting with T5-small for both IC and DST. When more labeled data are used or the model size is increased, improvement margins get smaller. 
The only exception is the 5-shot margin of \textsc{CIns} over PE for T5-base on DST, because this is the only case when PE underperforms STD.
Therefore, we contend that \textsc{CIns} is especially beneficial for low-resource learning with reasonable-sized models.

\section{Conclusion}

We study how to instruct PLMs for few-shot learning in ToD. \textsc{CIns} is proposed to augment state-of-the-art prompting techniques with extra task-specific definition and constraint.
Extensive empirical results on three ToD tasks demonstrate the consistent improvements of \textsc{CIns}. Our findings on using instructions may inspire future studies towards better utilizing PLMs for building more sample-efficient and scalable ToD systems.

\bibliography{CI}

\begin{thebibliography}{46}
\expandafter\ifx\csname natexlab\endcsname\relax\def\natexlab#1{#1}\fi

\bibitem[{Brown et~al.(2020)Brown, Mann, Ryder, Subbiah, Kaplan, Dhariwal,
  Neelakantan, Shyam, Sastry, Askell, Agarwal, Herbert{-}Voss, Krueger,
  Henighan, Child, Ramesh, Ziegler, Wu, Winter, Hesse, Chen, Sigler, Litwin,
  Gray, Chess, Clark, Berner, McCandlish, Radford, Sutskever, and
  Amodei}]{brown2020language}
Tom~B. Brown, Benjamin Mann, Nick Ryder, Melanie Subbiah, Jared Kaplan,
  Prafulla Dhariwal, Arvind Neelakantan, Pranav Shyam, Girish Sastry, Amanda
  Askell, Sandhini Agarwal, Ariel Herbert{-}Voss, Gretchen Krueger, Tom
  Henighan, Rewon Child, Aditya Ramesh, Daniel~M. Ziegler, Jeffrey Wu, Clemens
  Winter, Christopher Hesse, Mark Chen, Eric Sigler, Mateusz Litwin, Scott
  Gray, Benjamin Chess, Jack Clark, Christopher Berner, Sam McCandlish, Alec
  Radford, Ilya Sutskever, and Dario Amodei. 2020.
\newblock Language models are few-shot learners.
\newblock \emph{CoRR}, abs/2005.14165.

\bibitem[{Budzianowski and Vulic(2019)}]{budzianowski2019hello}
Pawel Budzianowski and Ivan Vulic. 2019.
\newblock Hello, it's {GPT-2} - how can {I} help you? towards the use of
  pretrained language models for task-oriented dialogue systems.
\newblock In \emph{NGT@EMNLP-IJCNLP}, pages 15--22. Association for
  Computational Linguistics.

\bibitem[{Budzianowski et~al.(2018)Budzianowski, Wen, Tseng, Casanueva, Ultes,
  Ramadan, and Gasic}]{budzianowski2018multiwoz}
Pawel Budzianowski, Tsung{-}Hsien Wen, Bo{-}Hsiang Tseng, I{\~{n}}igo
  Casanueva, Stefan Ultes, Osman Ramadan, and Milica Gasic. 2018.
\newblock Multiwoz - {A} large-scale multi-domain wizard-of-oz dataset for
  task-oriented dialogue modelling.
\newblock In \emph{{EMNLP}}, pages 5016--5026. Association for Computational
  Linguistics.

\bibitem[{Cai et~al.(2021)Cai, Zhou, Mi, and Faltings}]{cai2021slim}
Fengyu Cai, Wanhao Zhou, Fei Mi, and Boi Faltings. 2021.
\newblock Slim: Explicit slot-intent mapping with bert for joint multi-intent
  detection and slot filling.
\newblock \emph{arXiv preprint arXiv:2108.11711}.

\bibitem[{Chao and Lane(2019)}]{chao2019bert}
Guan-Lin Chao and Ian Lane. 2019.
\newblock Bert-dst: Scalable end-to-end dialogue state tracking with
  bidirectional encoder representations from transformer.
\newblock \emph{Proc. Interspeech 2019}, pages 1468--1472.

\bibitem[{Chen et~al.(2019)Chen, Zhuo, and Wang}]{chen2019bert}
Qian Chen, Zhu Zhuo, and Wen Wang. 2019.
\newblock Bert for joint intent classification and slot filling.
\newblock \emph{arXiv preprint arXiv:1902.10909}.

\bibitem[{Devlin et~al.(2019)Devlin, Chang, Lee, and
  Toutanova}]{devlin2018bert}
Jacob Devlin, Ming{-}Wei Chang, Kenton Lee, and Kristina Toutanova. 2019.
\newblock {BERT:} pre-training of deep bidirectional transformers for language
  understanding.
\newblock In \emph{{NAACL-HLT} {(1)}}, pages 4171--4186. Association for
  Computational Linguistics.

\bibitem[{Dong et~al.(2019)Dong, Yang, Wang, Wei, Liu, Wang, Gao, Zhou, and
  Hon}]{DBLP:conf/nips/00040WWLWGZH19}
Li~Dong, Nan Yang, Wenhui Wang, Furu Wei, Xiaodong Liu, Yu~Wang, Jianfeng Gao,
  Ming Zhou, and Hsiao{-}Wuen Hon. 2019.
\newblock Unified language model pre-training for natural language
  understanding and generation.
\newblock In \emph{NeurIPS}, pages 13042--13054.

\bibitem[{Du{\v{s}}ek and Jurcicek(2019)}]{duvsek2019neural}
Ond{\v{r}}ej Du{\v{s}}ek and Filip Jurcicek. 2019.
\newblock Neural generation for czech: Data and baselines.
\newblock In \emph{Proceedings of the 12th International Conference on Natural
  Language Generation}, pages 563--574.

\bibitem[{Gao et~al.(2021)Gao, Fisch, and Chen}]{DBLP:conf/acl/GaoFC20}
Tianyu Gao, Adam Fisch, and Danqi Chen. 2021.
\newblock Making pre-trained language models better few-shot learners.
\newblock In \emph{{ACL/IJCNLP} {(1)}}, pages 3816--3830. Association for
  Computational Linguistics.

\bibitem[{Ham et~al.(2020)Ham, Lee, Jang, and Kim}]{ham2020end}
DongHoon Ham, Jeong{-}Gwan Lee, Youngsoo Jang, and Kee{-}Eung Kim. 2020.
\newblock End-to-end neural pipeline for goal-oriented dialogue systems using
  {GPT-2}.
\newblock In \emph{{ACL}}, pages 583--592. Association for Computational
  Linguistics.

\bibitem[{Hosseini{-}Asl et~al.(2020)Hosseini{-}Asl, McCann, Wu, Yavuz, and
  Socher}]{hosseini2020simple}
Ehsan Hosseini{-}Asl, Bryan McCann, Chien{-}Sheng Wu, Semih Yavuz, and Richard
  Socher. 2020.
\newblock A simple language model for task-oriented dialogue.
\newblock \emph{CoRR}, abs/2005.00796.

\bibitem[{Hu et~al.(2021)Hu, Ding, Wang, Liu, Li, and
  Sun}]{DBLP:journals/corr/abs-2108-02035}
Shengding Hu, Ning Ding, Huadong Wang, Zhiyuan Liu, Juanzi Li, and Maosong Sun.
  2021.
\newblock Knowledgeable prompt-tuning: Incorporating knowledge into prompt
  verbalizer for text classification.
\newblock \emph{CoRR}, abs/2108.02035.

\bibitem[{Jiang et~al.(2020)Jiang, Xu, Araki, and
  Neubig}]{DBLP:journals/tacl/JiangXAN20}
Zhengbao Jiang, Frank~F. Xu, Jun Araki, and Graham Neubig. 2020.
\newblock How can we know what language models know.
\newblock \emph{Trans. Assoc. Comput. Linguistics}, 8:423--438.

\bibitem[{Kale and Rastogi(2020{\natexlab{a}})}]{kale-rastogi-2020-template}
Mihir Kale and Abhinav Rastogi. 2020{\natexlab{a}}.
\newblock Template guided text generation for task oriented dialogue.
\newblock In \emph{Proceedings of the 2020 Conference on Empirical Methods in
  Natural Language Processing (EMNLP)}, pages 6505--6520, Online. Association
  for Computational Linguistics.

\bibitem[{Kale and Rastogi(2020{\natexlab{b}})}]{kale2020text}
Mihir Kale and Abhinav Rastogi. 2020{\natexlab{b}}.
\newblock Text-to-text pre-training for data-to-text tasks.
\newblock In \emph{Proceedings of the 13th International Conference on Natural
  Language Generation}, pages 97--102.

\bibitem[{Larson et~al.(2019)Larson, Mahendran, Peper, Clarke, Lee, Hill,
  Kummerfeld, Leach, Laurenzano, Tang, and Mars}]{larson2019evaluation}
Stefan Larson, Anish Mahendran, Joseph~J. Peper, Christopher Clarke, Andrew
  Lee, Parker Hill, Jonathan~K. Kummerfeld, Kevin Leach, Michael~A. Laurenzano,
  Lingjia Tang, and Jason Mars. 2019.
\newblock An evaluation dataset for intent classification and out-of-scope
  prediction.
\newblock In \emph{{EMNLP/IJCNLP} {(1)}}, pages 1311--1316. Association for
  Computational Linguistics.

\bibitem[{Lester et~al.(2021)Lester, Al{-}Rfou, and
  Constant}]{DBLP:journals/corr/abs-2104-08691}
Brian Lester, Rami Al{-}Rfou, and Noah Constant. 2021.
\newblock The power of scale for parameter-efficient prompt tuning.
\newblock \emph{CoRR}, abs/2104.08691.

\bibitem[{Lewis et~al.(2019)Lewis, Liu, Goyal, Ghazvininejad, Mohamed, Levy,
  Stoyanov, and Zettlemoyer}]{lewis2019bart}
Mike Lewis, Yinhan Liu, Naman Goyal, Marjan Ghazvininejad, Abdelrahman Mohamed,
  Omer Levy, Ves Stoyanov, and Luke Zettlemoyer. 2019.
\newblock Bart: Denoising sequence-to-sequence pre-training for natural
  language generation, translation, and comprehension.
\newblock \emph{arXiv preprint arXiv:1910.13461}.

\bibitem[{Li and Liang(2021)}]{DBLP:conf/acl/LiL20}
Xiang~Lisa Li and Percy Liang. 2021.
\newblock Prefix-tuning: Optimizing continuous prompts for generation.
\newblock In \emph{{ACL/IJCNLP} {(1)}}, pages 4582--4597. Association for
  Computational Linguistics.

\bibitem[{Lin et~al.(2021)Lin, Liu, Moon, Crook, Zhou, Wang, Yu, Madotto, Cho,
  and Subba}]{lin2021leveraging}
Zhaojiang Lin, Bing Liu, Seungwhan Moon, Paul~A Crook, Zhenpeng Zhou, Zhiguang
  Wang, Zhou Yu, Andrea Madotto, Eunjoon Cho, and Rajen Subba. 2021.
\newblock Leveraging slot descriptions for zero-shot cross-domain dialogue
  statetracking.
\newblock In \emph{{NAACL-HLT} {(1)}}, pages 5640--5648. Association for
  Computational Linguistics.

\bibitem[{Liu et~al.(2021{\natexlab{a}})Liu, Shen, Zhang, Dolan, Carin, and
  Chen}]{DBLP:journals/corr/abs-2101-06804}
Jiachang Liu, Dinghan Shen, Yizhe Zhang, Bill Dolan, Lawrence Carin, and Weizhu
  Chen. 2021{\natexlab{a}}.
\newblock What makes good in-context examples for gpt-3?
\newblock \emph{CoRR}, abs/2101.06804.

\bibitem[{Liu et~al.(2021{\natexlab{b}})Liu, Yuan, Fu, Jiang, Hayashi, and
  Neubig}]{DBLP:journals/corr/abs-2107-13586}
Pengfei Liu, Weizhe Yuan, Jinlan Fu, Zhengbao Jiang, Hiroaki Hayashi, and
  Graham Neubig. 2021{\natexlab{b}}.
\newblock Pre-train, prompt, and predict: {A} systematic survey of prompting
  methods in natural language processing.
\newblock \emph{CoRR}, abs/2107.13586.

\bibitem[{Liu et~al.(2021{\natexlab{c}})Liu, Zheng, Du, Ding, Qian, Yang, and
  Tang}]{DBLP:journals/corr/abs-2103-10385}
Xiao Liu, Yanan Zheng, Zhengxiao Du, Ming Ding, Yujie Qian, Zhilin Yang, and
  Jie Tang. 2021{\natexlab{c}}.
\newblock {GPT} understands, too.
\newblock \emph{CoRR}, abs/2103.10385.

\bibitem[{Mi et~al.(2020)Mi, Chen, Zhao, Huang, and Faltings}]{mi2020continual}
Fei Mi, Liangwei Chen, Mengjie Zhao, Minlie Huang, and Boi Faltings. 2020.
\newblock Continual learning for natural language generation in task-oriented
  dialog systems.
\newblock In \emph{Findings of the Association for Computational Linguistics:
  EMNLP 2020}, pages 3461--3474.

\bibitem[{Mi et~al.(2019)Mi, Huang, Zhang, and Faltings}]{mi2019meta}
Fei Mi, Minlie Huang, Jiyong Zhang, and Boi Faltings. 2019.
\newblock Meta-learning for low-resource natural language generation in
  task-oriented dialogue systems.
\newblock In \emph{Proceedings of the 28th International Joint Conference on
  Artificial Intelligence}, pages 3151--3157.

\bibitem[{Mi et~al.(2021)Mi, Zhou, Kong, Cai, Huang, and Faltings}]{mi2021self}
Fei Mi, Wanhao Zhou, Lingjing Kong, Fengyu Cai, Minlie Huang, and Boi Faltings.
  2021.
\newblock Self-training improves pre-training for few-shot learning in
  task-oriented dialog systems.
\newblock In \emph{Proceedings of the 2021 Conference on Empirical Methods in
  Natural Language Processing}, pages 1887--1898.

\bibitem[{Mishra et~al.(2021)Mishra, Khashabi, Baral, and
  Hajishirzi}]{DBLP:journals/corr/abs-2104-08773}
Swaroop Mishra, Daniel Khashabi, Chitta Baral, and Hannaneh Hajishirzi. 2021.
\newblock Natural instructions: Benchmarking generalization to new tasks from
  natural language instructions.
\newblock \emph{CoRR}, abs/2104.08773.

\bibitem[{Papineni et~al.(2002)Papineni, Roukos, Ward, and
  Zhu}]{papineni2002bleu}
Kishore Papineni, Salim Roukos, Todd Ward, and Wei-Jing Zhu. 2002.
\newblock Bleu: a method for automatic evaluation of machine translation.
\newblock In \emph{Proceedings of the 40th annual meeting of the Association
  for Computational Linguistics}, pages 311--318.

\bibitem[{Peng et~al.(2020{\natexlab{a}})Peng, Li, Li, Shayandeh, Liden, and
  Gao}]{peng2020soloist}
Baolin Peng, Chunyuan Li, Jinchao Li, Shahin Shayandeh, Lars Liden, and
  Jianfeng Gao. 2020{\natexlab{a}}.
\newblock {SOLOIST:} few-shot task-oriented dialog with {A} single pre-trained
  auto-regressive model.
\newblock \emph{CoRR}, abs/2005.05298.

\bibitem[{Peng et~al.(2020{\natexlab{b}})Peng, Zhu, Li, Li, Li, Zeng, and
  Gao}]{peng2020few}
Baolin Peng, Chenguang Zhu, Chunyuan Li, Xiujun Li, Jinchao Li, Michael Zeng,
  and Jianfeng Gao. 2020{\natexlab{b}}.
\newblock Few-shot natural language generation for task-oriented dialog.
\newblock In \emph{{EMNLP} (Findings)}, pages 172--182. Association for
  Computational Linguistics.

\bibitem[{Qin and Eisner(2021)}]{DBLP:conf/naacl/QinE21}
Guanghui Qin and Jason Eisner. 2021.
\newblock Learning how to ask: Querying lms with mixtures of soft prompts.
\newblock In \emph{{NAACL-HLT}}, pages 5203--5212. Association for
  Computational Linguistics.

\bibitem[{Radford et~al.(2018)Radford, Narasimhan, Salimans, and
  Sutskever}]{radford2018improving}
Alec Radford, Karthik Narasimhan, Tim Salimans, and Ilya Sutskever. 2018.
\newblock Improving language understanding by generative pre-training.

\bibitem[{Radford et~al.(2019)Radford, Wu, Child, Luan, Amodei, and
  Sutskever}]{radford2019language}
Alec Radford, Jeffrey Wu, Rewon Child, David Luan, Dario Amodei, and Ilya
  Sutskever. 2019.
\newblock Language models are unsupervised multitask learners.
\newblock \emph{OpenAI blog}, 1(8):9.

\bibitem[{Raffel et~al.(2020)Raffel, Shazeer, Roberts, Lee, Narang, Matena,
  Zhou, Li, and Liu}]{raffel2020exploring}
Colin Raffel, Noam Shazeer, Adam Roberts, Katherine Lee, Sharan Narang, Michael
  Matena, Yanqi Zhou, Wei Li, and Peter~J Liu. 2020.
\newblock Exploring the limits of transfer learning with a unified text-to-text
  transformer.
\newblock \emph{Journal of Machine Learning Research}, 21:1--67.

\bibitem[{Rastogi et~al.(2019)Rastogi, Zang, Sunkara, Gupta, and
  Khaitan}]{rastogi2019scalable}
Abhinav Rastogi, Xiaoxue Zang, Srinivas Sunkara, Raghav Gupta, and Pranav
  Khaitan. 2019.
\newblock {Towards Scalable Multi-domain Conversational Agents: The
  Schema-Guided Dialogue Dataset}.
\newblock In \emph{{Proceedings of the AAAI Conference on Artificial
  Intelligence}}.

\bibitem[{Rastogi et~al.(2020)Rastogi, Zang, Sunkara, Gupta, and
  Khaitan}]{rastogi2020towards}
Abhinav Rastogi, Xiaoxue Zang, Srinivas Sunkara, Raghav Gupta, and Pranav
  Khaitan. 2020.
\newblock Towards scalable multi-domain conversational agents: The
  schema-guided dialogue dataset.
\newblock In \emph{Proceedings of the AAAI Conference on Artificial
  Intelligence}, volume~34, pages 8689--8696.

\bibitem[{Schick and Sch{\"{u}}tze(2020)}]{DBLP:journals/corr/abs-2012-11926}
Timo Schick and Hinrich Sch{\"{u}}tze. 2020.
\newblock Few-shot text generation with pattern-exploiting training.
\newblock \emph{CoRR}, abs/2012.11926.

\bibitem[{Schick and
  Sch{\"{u}}tze(2021{\natexlab{a}})}]{DBLP:conf/eacl/SchickS21}
Timo Schick and Hinrich Sch{\"{u}}tze. 2021{\natexlab{a}}.
\newblock Exploiting cloze-questions for few-shot text classification and
  natural language inference.
\newblock In \emph{{EACL}}, pages 255--269. Association for Computational
  Linguistics.

\bibitem[{Schick and
  Sch{\"{u}}tze(2021{\natexlab{b}})}]{DBLP:conf/naacl/SchickS21}
Timo Schick and Hinrich Sch{\"{u}}tze. 2021{\natexlab{b}}.
\newblock It's not just size that matters: Small language models are also
  few-shot learners.
\newblock In \emph{{NAACL-HLT}}, pages 2339--2352. Association for
  Computational Linguistics.

\bibitem[{Tam et~al.(2021)Tam, Menon, Bansal, Srivastava, and
  Raffel}]{DBLP:journals/corr/abs-2103-11955}
Derek Tam, Rakesh~R. Menon, Mohit Bansal, Shashank Srivastava, and Colin
  Raffel. 2021.
\newblock Improving and simplifying pattern exploiting training.
\newblock \emph{CoRR}, abs/2103.11955.

\bibitem[{Wen et~al.(2015)Wen, Gasic, Mrk{\v{s}}i{\'c}, Su, Vandyke, and
  Young}]{wen2015semantically}
Tsung-Hsien Wen, Milica Gasic, Nikola Mrk{\v{s}}i{\'c}, Pei-Hao Su, David
  Vandyke, and Steve Young. 2015.
\newblock Semantically conditioned lstm-based natural language generation for
  spoken dialogue systems.
\newblock In \emph{Proceedings of the 2015 Conference on Empirical Methods in
  Natural Language Processing}, pages 1711--1721.

\bibitem[{Williams and Zipser(1989)}]{williams1989learning}
Ronald~J Williams and David Zipser. 1989.
\newblock A learning algorithm for continually running fully recurrent neural
  networks.
\newblock \emph{Neural Computation}, 1(2):270--280.

\bibitem[{Wu et~al.(2020)Wu, Hoi, Socher, and Xiong}]{wu2020tod}
Chien-Sheng Wu, Steven Hoi, Richard Socher, and Caiming Xiong. 2020.
\newblock {TOD-BERT}: Pre-trained natural language understanding for
  task-oriented dialogues.
\newblock In \emph{EMNLP}, pages 917--929. Association for Computational
  Linguistics.

\bibitem[{Wu et~al.(2019)Wu, Madotto, Hosseini{-}Asl, Xiong, Socher, and
  Fung}]{wu2019transferable}
Chien{-}Sheng Wu, Andrea Madotto, Ehsan Hosseini{-}Asl, Caiming Xiong, Richard
  Socher, and Pascale Fung. 2019.
\newblock Transferable multi-domain state generator for task-oriented dialogue
  systems.
\newblock In \emph{{ACL} {(1)}}, pages 808--819. Association for Computational
  Linguistics.

\bibitem[{Zhang et~al.(2020)Zhang, Hashimoto, Wu, Wang, Philip, Socher, and
  Xiong}]{zhang2020find}
Jianguo Zhang, Kazuma Hashimoto, Chien-Sheng Wu, Yao Wang, S~Yu Philip, Richard
  Socher, and Caiming Xiong. 2020.
\newblock Find or classify? dual strategy for slot-value predictions on
  multi-domain dialog state tracking.
\newblock In \emph{Proceedings of the Ninth Joint Conference on Lexical and
  Computational Semantics}, pages 154--167.

\end{thebibliography}

\bibliographystyle{acl_natbib}

\newpage
~\newpage

\appendix

\part*{ Appendix}

\section{Reproducibility Checklist}

\subsection{Dataset Specifications for Different Tasks}
\label{appendix:data}

In table \ref{table:dataset}, we report the exact training and testing dataset statistics in different tasks and configurations. In few-shot settings, the size of validation data is set the same as training data.

\begin{table}[htb!]
\small
    \centering
    \setlength{\tabcolsep}{1pt}
    \begin{tabular}{c|c|c}
        \toprule
          \multirow{4}{*}{\textbf{OOS}} & 1-shot Train & 15 \\ 
           & 5-shot Train & 75 \\ 
           & Full Train & 1500 \\
           & Test & 450 \\
         \midrule
         \multirow{4}{*}{\textbf{MultiWoZ2.0}} & 1\% Train & (28, 34, 37, 11, 30) \\ 
           & 5\% Train & (129, 168, 197, 80, 147) \\
           & Full Train & (2717, 3381, 3813, 1654, 3103) \\
           & Test & (395, 384, 437, 195, 494) \\
          \midrule
         \multirow{4}{*}{\textbf{FewShotSGD}} & 5-shot Train & 558   \\ 
           & 10-shot Train & 1,075 \\ 
           & Full Train &  164,978 \\
           & Test & 42,297 \\
         
        \bottomrule
    \end{tabular}
    \caption{Training and testing dataset statistics. In few-shot settings. The number of utterances are reported for OOS and FewShotSDG, and the number of dialogs are reported for MultiWoZ2.0. For OOS, statistics corresponds to a single domain, and all domain statistics are the same. For MultiWoZ2.0, the reported statistics are average of three random sampling seeds, and it corresponds to 5 domains in the order of (Attraction, Hotel, Restaurant, Taxi, Train). }
    \label{table:dataset}
    \vspace{-0.1in}
\end{table}

\begin{figure*}[t!]
    \centering
    \includegraphics[width=1\textwidth]{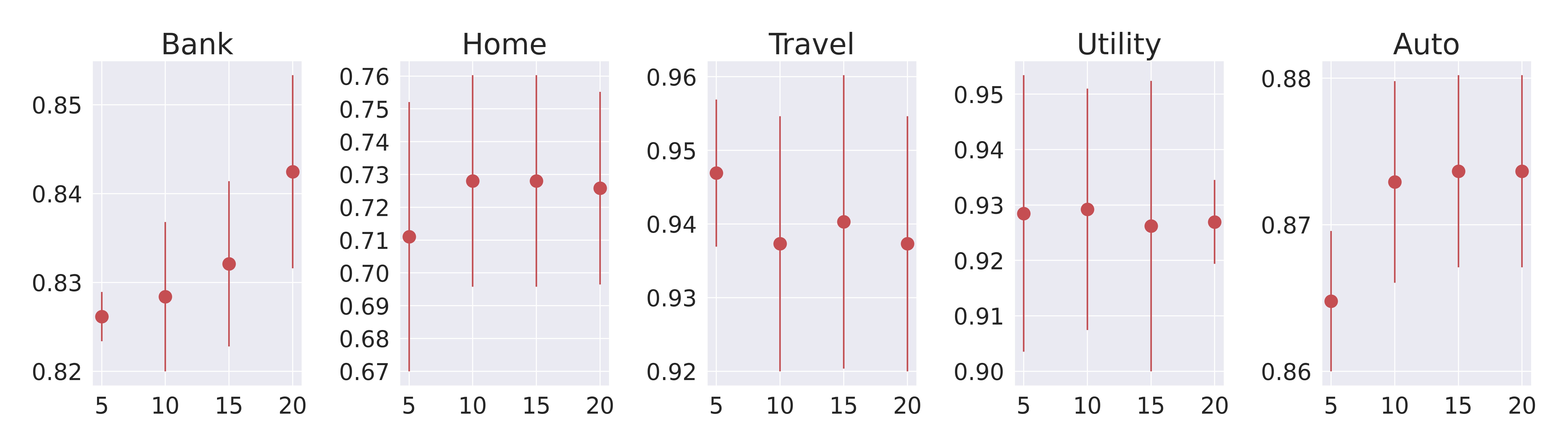}
    \caption{Accuracy with standard deviation of \textsc{CIns} in different domains with varying number (5,10,15,20) of validation data per intent label for the 5-shot intent classification task. Validation data are only used for early-stopping T5-small model training. Validation size 5 is the same as the 5-shot training data.}
    \label{fig:devsize}
\end{figure*}

\subsection{Hyper-parameter Settings}
\label{appendix:hyper}

As we mentioned in our main paper \S \ref{sec:exp_setting}, we train T5-small and T5-base with batch size 8 for 30 epochs for IC, 20 epochs for DST, and 50 epochs for NLG.  
The maximum sequence length for T5 encoder is set to 128 for NLG, 256 for IC, and 512 for DST as the input contains dialog history.
All models are trained using AdamW optimizer with the initial learning rate of 1e-4 for DST and NLG, and 3e-4 for IC. Other parameters are set by default for AdamW, and we only tune the initial learning rate within \{1e-4, 3e-4, 5e-4\}.

\subsection{Experiment Prompts}
\label{appendix:prompts}

As we mentioned in our main paper \S \ref{sec:exp_setting}, we tested four different prompts for both PE and \textsc{CIns}, and report the performance of the best one. The exact prompts tested are included in Table \ref{table:exp_prompts}, and the ones reported for PE and \textsc{CIns} are highlighted in the last column.
For ``Naive'' representation in NLG, question prompt expressions do not necessarily outperforms short declarative expressions. This observation is different from other tasks and settings reported in our main paper Figure \ref{fig:prompt_type}. Nevertheless, the performance of ``Naive'' lags far behind ``T2G2'' for NLG.

\subsection{Intent Descriptions for Intent Classification}
\label{appendix:ic_description}

As we explained in our main paper (c.f. Table \ref{table:cp_def} and \S \ref{sec:ci_tod}), the constraint component of \textsc{CIns} for intent classification task contains label/intent descriptions. In Table \ref{table:intent desc1} and \ref{table:intent desc2}, we provide a detailed documentation of intents in different domains and their corresponding short descriptions used in our experiments. 

\section{Supplementary Results}

\subsection{Effect of Validation Data Size}
\label{appendix:val_size}

As mentioned in our main paper, we use the validation data the same size as the few-shot training data in different ToD tasks.
A large number of validation data contradict realistic few-shot learning scenarios. 
In this experiment, we test the effect of different validation data size w.r.t. model performance in Figure \ref{fig:devsize}. As a case study, we provide the results of Comprehensive Instruction (\textsc{CIns}) in different domains with a varying number (5,10,15,20) of validation data per intent label for the 5-shot intent classification task. Validation size 5 is the same as the 5-shot training. Results reported in Figure \ref{fig:devsize} only use validation data for early-stopping T5-small model training without being used for tuning other hyper-parameters.

We could see that the performance varies with the validation data size. Better accuracy is achieved with larger validation data size for domains except for ``Travel''. In this experiment, we aim to demonstrate the sensitivity to validation, but it does not mean that larger validation data achieve better performance. 
As the distribution similarity between validation and test data determines the effectiveness of the validation data, the size of validation data could be a factor influencing the final performance.
More in-depth explorations regarding the choice of validation data or even few-shot training data are left as future work.

\begin{table*}[htb!]
    \centering
    \begin{tabular}{c|l|c}
        \toprule
        \multirow{6}{*}{\textbf{IC}} &  The given query asks about: & \\
        & Question: What does the given query ask about? & \\
        & The topic of the given query is: & \\
        & Question: What is the topic of the given query? & \\
        & The intent of the given query is: & \\
        & Question: What is the intent of the given query? &  PE \& \textsc{CIns} \\
        \midrule
        \multirowcell{6}{\textbf{DST} \\ (hotel\_stars)} &  Stars of the hotel: & \\
        & Question: What is the stars of the hotel? & \\
        & Star rating of the hotel: $^\dag$ & \\
        & Question: What is star rating of the hotel? $^\dag$ & \\
        & Number of stars of the hotel: & \\
        & Question: What is the number of stars of the hotel? & PE \& \textsc{CIns} \\
        \midrule
        \multirowcell{6}{\textbf{NLG} \\ (Naive)} &  Verbalize the input: & PE \& \textsc{CIns} \\
        & Question: what is the verbalized utterance of the input? &  \\
        & Question: how to verbalize the input? & \\
        & Paraphrase the input: & \\
        & Question: how to paraphrase the input? & \\
        & Question: what is the paraphrase of the input? &  \\
        \midrule
        \multirowcell{6}{\textbf{NLG} \\ (T2G2)} &  Rewrite the input: & \\
        & Question: what is the rewriting of the input? & PE \\
        & Question: how to rewrite the input? & \\
        & Paraphrase the input: & \\
        & Question: how to paraphrase the input? & \\
        & Question: what is the paraphrase of the input? & \textsc{CIns} \\
        \bottomrule
    \end{tabular}
    \caption{Different prompt tested for IC, DST and NLG. For DST, we use the slot ``hotel\_stars'' as an example. $^\dag$ it corresponds to human-written slot description in \citet{lin2021leveraging}.}
    \label{table:exp_prompts}
    \vspace{-0.1in}
\end{table*}

\begin{table*}[t!]
    \centering
    \begin{tabular}{c|l|l}
        \toprule
        \textbf{Domain} & \textbf{Intent} & \textbf{Intent Description} \\
        \midrule
        \multirow{15}{*}{\textbf{Bank}} &  freeze account   &   stop, freeze, or block a bank account  \\ 
       &   routing   &   ask bank account routing number  \\ 
       &   pin change   &   query or operate bank account pin number  \\ 
       &   bill due   &   due date about bill  \\ 
       &   pay bill   &   pay bill information  \\ 
       &   account blocked   &   why account is blocked  \\ 
       &   interest rate   &   ask information about interest rate  \\ 
       &   min payment   &   ask mimnimum, least, or lowest amount to pay  \\ 
       &   bill balance   &   ask bill balance  \\ 
       &   transfer   &   transfer money between account  \\ 
       &   order checks   &   order new checks  \\ 
       &   balance   &   ask balance of a bank account  \\ 
       &   spending history   &   ask spend history  \\ 
       &   transactions   &   ask the last or recent transactions  \\ 
       &   report fraud   &   report fraud or wrong transaction \\
       \midrule
       \multirow{15}{*}{\textbf{Home}} & what song & ask name of a song \\ 
        & play music & play or hear a song \\ 
        & todo list update & add, remove, cross, or update event on todo list \\ 
        & reminder & remind, recall things to do \\ 
        & reminder update & create, make, or set a reminder \\ 
        & calendar update & add or remove a calender event \\ 
        & order status & ask order delivery information \\ 
        & update playlist & add song to playlist \\ 
        & shopping list & ask information about shopping list \\ 
        & calendar & ask about event on calender \\ 
        & next song & play the next song \\ 
        & order & make an order to buy something \\ 
        & todo list & ask information about todo list \\ 
        & shopping list update & add or remove item on shopping list \\ 
        & smart home & control tv, ac, oven, light \\
       \midrule
       \multirow{15}{*}{\textbf{Travel}} & plug type & information about plug or socket type converter \\ 
        & travel notification & tell, inform, alert bank about travel \\ 
        & translate & ask translation of an expression \\ 
        & flight status & ask flight status \\ 
        & international visa & ask international visa requirement to travel \\ 
        & timezone & ask timezone of a place \\ 
        & exchange rate & ask exchange rate between two currency \\ 
        & travel suggestion & travel suggestion on activities, attractions or things to do \\ 
        & travel alert & ask about travel alert or safety \\ 
        & vaccines & vaccine shots to travel \\ 
        & lost luggage & ask airlines about lost luggage during flight \\ 
        & book flight & ask, find or book a flight \\ 
        & book hotel & ask, find or book a hotel \\ 
        & carry on & ask information about carry on rule \\ 
        & car rental & book or rent a car \\
        \bottomrule
    \end{tabular}
    \caption{Intents and intent descriptions in different domains in our experiments. Intent descriptions are used in our CI constraints.}
    \label{table:intent desc1}
    \vspace{-0.1in}
\end{table*}

\begin{table*}[t!]
    \centering
    \begin{tabular}{c|l|l}
        \toprule
        \textbf{Domain} & \textbf{Intent} & \textbf{Intent Description} \\
        \midrule
       \multirow{15}{*}{\textbf{Utility}} & weather & temperature or weather forecast \\ 
        & alarm & set an alarm \\ 
        & date & ask about date \\ 
        & find phone & where to find lost cellphone \\ 
        & share location & share or send location \\ 
        & timer & set a reminder timer \\ 
        & make call & action about phone call or dial \\ 
        & calculator & calculate math: plus, minus, multiple, divide, square root etc \\ 
        & definition & ask definition or meaning \\ 
        & measurement conversion & conversion between two measurements \\ 
        & flip coin & flip a coin \\ 
        & spelling & ask or correct spelling \\ 
        & time & ask about time \\ 
        & roll dice & roll a dice with several sides \\ 
        & text & text a message \\
       \midrule
       \multirow{15}{*}{\textbf{Auto}} & current location & ask current location \\ 
        & oil change when & when is due for next oil change \\ 
        & oil change how & instructions how to change oil \\ 
        & uber & book, find, get an uber \\ 
        & traffic & ask traffic situation \\ 
        & tire pressure & ask or check car tire pressure \\ 
        & schedule maintenance & book, schedule or appoint a maintenance for car \\ 
        & gas & ask gas tank or fuel \\ 
        & mpg & ask gas mileage mpg \\ 
        & distance & ask travel time or distance \\ 
        & directions & ask directions to a destination \\ 
        & last maintenance & ask about last maintenance or service \\ 
        & gas type & ask gas type of the car \\ 
        & tire change & when to change or replace tire \\ 
        & jump start & jump start a battery \\
        \bottomrule
    \end{tabular}
    \caption{Continuation of Table \ref{table:intent desc1}}
    \label{table:intent desc2}
    \vspace{-0.1in}
\end{table*}

\end{document}